\documentclass[letterpaper]{article} 
\usepackage[preprint]{aaai2027}  
\usepackage[hyphens]{url}  
\usepackage{graphicx} 
\usepackage{natbib}  
\usepackage{caption} 
\usepackage{algorithm}
\usepackage{algorithmic}
\usepackage{bm}
\usepackage{tikz}
\usepackage{xcolor}
\usepackage{amssymb}
\usepackage{amsmath}
\usepackage{amsfonts}
\usepackage{multirow}
\usepackage{subcaption}
\usepackage[table]{xcolor} 
\usepackage{makecell}

\usepackage{newfloat}
\usepackage{listings}
\DeclareCaptionStyle{ruled}{labelfont=normalfont,labelsep=colon,strut=off} 
\floatstyle{ruled}
\newfloat{listing}{tb}{lst}{}
\floatname{listing}{Listing}

\usepackage{booktabs}

\title{RSRA: Training-Free Probing of Representation Sensitivity for Efficient LoRA Rank Allocation\thanks{Preprint.}}
\author{
    Jiaqi Liu\textsuperscript{\rm 1},
    Haidong Kang\textsuperscript{\rm 2}\corresponding,
    Qihui Zhao\textsuperscript{\rm2},
    Guo Yu\textsuperscript{\rm3},
    Jingchao Wang\textsuperscript{\rm4}
}

\affiliations{
    \textsuperscript{\rm 1}Dalian University of Technology\\
    \textsuperscript{\rm 2}Northeastern University\\
    \textsuperscript{\rm 3}Nanjing Tech University\\
    \textsuperscript{\rm 4}Peking University\\
}

\begin{document}

\maketitle

\begin{abstract}
Parameter-efficient fine-tuning enables large language models to adapt to downstream tasks with substantially lower computational and storage cost, and Low-Rank Adaptation (LoRA) is among its most widely used techniques. However, vanilla LoRA assigns a uniform rank to all adapted modules, while existing adaptive methods either incur additional optimization overhead or rely on static weights and local gradients that do not capture task-conditioned representation changes. We propose RSRA, a training-free rank allocator that estimates where adaptation capacity is most needed through forward-only representation sensitivity probing on a small calibration set. Specifically, RSRA uses Spectral Effective Rank to allocate capacity across layers, measures module-wise hidden-state displacement under standardized virtual low-rank updates with the Fr\'{e}chet Distance, and combines both signals through hierarchical normalization to produce a task-aware rank configuration before fine-tuning. Across commonsense reasoning and natural language understanding benchmarks with Qwen3-4B and Mistral-7B, RSRA achieves the highest average performance in all three reported model--benchmark settings and a $1.48\times$--$1.93\times$ speedup in allocation time over the fastest competing pre-allocation method. When integrated with DoRA, LoRA-FA, and PiSSA, RSRA improves 15 of the 18 evaluated combinations and increases the average performance of all three PEFT methods.
\end{abstract}

\section{Introduction}
\label{sec:introduction}
\begin{figure}[t]
    \raggedright
    \includegraphics[
        width=0.9\columnwidth,
    ]{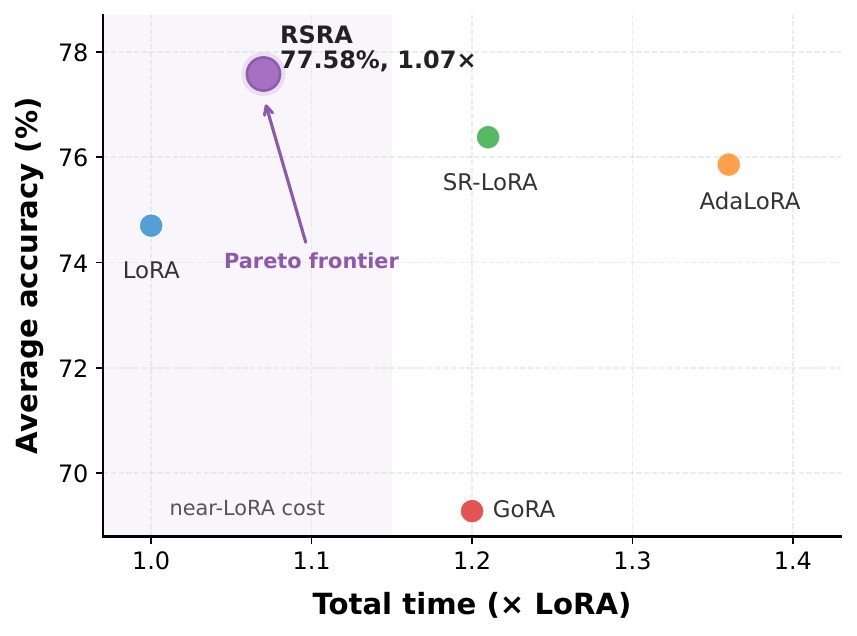}
    \caption{
    Performance--efficiency trade-off on GLUE with Mistral-7B.RSRA achieves the highest average accuracy while requiring only $1.07\times$ the total runtime of LoRA, yielding a Pareto-optimal trade-off among the compared methods.
    }
    \label{fig:intro}
\end{figure}

Large language models (LLMs) have demonstrated strong capabilities across a broad range of natural language processing tasks, yet adapting them to a specific domain or downstream objective through full-parameter fine-tuning remains computationally and storage intensive~\citep{brown2020language,ouyang2022training}. 
Parameter-Efficient Fine-Tuning (PEFT) addresses this challenge by updating only a small subset of parameters while retaining most of the knowledge encoded in the pretrained model~\citep{houlsby2019parameter,he2021towards}. 
Among existing PEFT techniques, Low-Rank Adaptation (LoRA) has become a widely adopted approach because it represents weight updates using compact low-rank matrices without introducing additional inference latency~\citep{hu2022lora}. 
However, vanilla LoRA typically assigns the same rank to every adapted module, implicitly assuming that different layers and module types require equal adaptation capacity. 
This assumption has motivated adaptive rank-allocation methods such as AdaLoRA~\citep{zhang2023adalora}, SoRA~\citep{ding2023sora}, ALoRA~\citep{liu2024alora}, SR-LoRA~\citep{zhang2025srlora}, and GoRA\citep{he2025gora}, which redistribute a limited parameter budget according to optimization dynamics, pretrained weight statistics, or local gradient information.

Despite this progress, determining an informative and computationally efficient allocation signal remains an open problem. 
Training-time allocation methods can dynamically identify important components, but often introduce iterative pruning, regularization, or structural adjustment during optimization. 
In contrast, existing pre-allocation approaches reduce training overhead but commonly rely on static properties or local first-order signals, which may not describe how an update changes task-conditioned representations as information propagates through the network. 
In Section~\ref{sec:motivation}, we investigate this problem from a representation perspective and observe that low-rank updates with the same rank induce substantially different changes in hidden-state distributions across both network depth and module type. 
Moreover, removing modules with higher representation sensitivity causes considerably greater downstream degradation than removing randomly selected or low-sensitivity modules. 
These observations reveal a hierarchical sensitivity pattern: adaptation requirements differ not only across layers, but also among modules within the same layer. 
They therefore lead to the central question of this work: \emph{Can task-relevant representation sensitivity be estimated before fine-tuning, without gradients, auxiliary training, or expensive rank search?}

To address this question, we propose \textbf{Representation Sensitivity Rank Allocation (RSRA)}, a training-free and  forward-only rank allocator that estimates representation sensitivity using a small task-specific calibration set. 
As detailed in Section~\ref{sec:methods}, RSRA measures the spectral breadth of each layer's hidden representations using Spectral Effective Rank, providing an inter-layer estimate of how broadly information is distributed across representation directions. 
It then applies standardized virtual low-rank probing to individual modules and measures the resulting displacement of the corresponding hidden-state distribution using the Fr\'echet Distance. 
The two signals are combined through hierarchical normalization to distribute a prescribed average-rank budget across layers and modules. 
The virtual probing is removed after allocation, and the resulting rank configuration can be used with the original fine-tuning and inference procedures. 
Consequently, RSRA operates only at the rank-allocation level and is broadly compatible with respect to the underlying adapter parameterization, initialization, and optimization strategy. 
This property makes it orthogonal to other PEFT variants and enables it to serve as a plug-and-play allocator for methods such as DoRA~\citep{liu2024dora}, LoRA-FA~\citep{zhang2023lora_fa}, and PiSSA~\citep{meng2024pissa}.

Extensive experiments demonstrate that RSRA provides a favorable balance between downstream performance and computational efficiency. 
As illustrated in Figure~\ref{fig:intro}, RSRA reaches the highest average GLUE\citep{wang2018glue} accuracy of $77.58\%$ on Mistral-7B\citep{jiang2023mistral} while requiring only $1.07\times$ the total time of LoRA, placing it on the performance--efficiency Pareto frontier among the compared methods. 
Across all three evaluated model--benchmark settings, RSRA achieves the highest average performance among LoRA, AdaLoRA, SR-LoRA, and GoRA, while its forward-only procedure also reduces standalone allocation time by $1.48\times$--$1.93\times$ relative to the fastest competing pre-allocation method. Beyond direct comparisons, integrating RSRA with DoRA, LoRA-FA, and PiSSA improves 15 of the 18 evaluated task--variant combinations and increases the average performance of all three PEFT methods. Module-removal experiments further show that modules assigned higher sensitivity scores make consistently larger functional contributions after fine-tuning, supporting the task relevance of the proposed probing signal.

Our main contributions are summarized as follows:
\begin{itemize}
    \item We identify a hierarchical pattern of representation sensitivity in LoRA adaptation: low-rank updates exhibit substantial heterogeneity across both network depth and module type, and modules with higher sensitivity contribute more strongly to downstream performance. This finding provides a representation-level motivation for adaptive rank allocation.

    \item We introduce RSRA, a training-free allocation procedure that combines inter-layer spectral breadth with intra-layer representation sensitivity estimated through standardized virtual low-rank probing. RSRA requires only forward passes on a small calibration set and is orthogonal to other PEFT variants, allowing it to be integrated without modifying their original adaptation mechanisms.


    \item We conduct extensive evaluations across commonsense reasoning and natural language understanding tasks. The results demonstrate consistent aggregate improvements over representative rank-allocation baselines, substantially reduced allocation time, broad compatibility with diverse PEFT methods, and functional alignment between the probing scores and trained-module contributions.
\end{itemize}

\begin{figure}[t]
    \centering
    \includegraphics[
    width=0.96\columnwidth
    ]{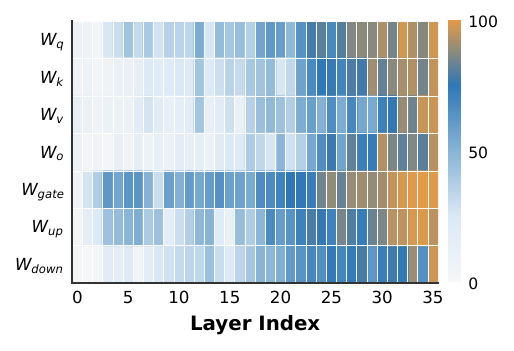}
    \caption{
    Heatmap of representation-sensitivity percentiles for a
    LoRA adapter. Sensitivity varies strongly across layers and module types.
    }
    \label{fig:boolq_fd_heatmap}
\end{figure}

\section{Motivation}
\label{sec:motivation}

Vanilla LoRA commonly assigns the same rank to all adapted modules.
This simple uniform design has motivated a growing line of adaptive
methods that redistribute a fixed rank budget across the network.
A central question for adaptive allocation is therefore:
\textbf{\emph{what signal should determine where additional rank capacity is placed?}}

We investigate this question from the representation perspective.
We define \emph{representation sensitivity} as the change in a layer's
task-conditioned hidden-state distribution caused by a module-specific
low-rank update. Operationally, it can be measured by comparing two model
states that differ only in the low-rank update associated with one module.
A highly sensitive module produces a large change in the layer
representations, whereas a less sensitive module produces only a limited
change.

We first study representation sensitivity in a standard LoRA adapter after fine-tuning. Specifically, we train vanilla LoRA on
BoolQ~\citep{clark2019boolq} and remove one learned module update at a time while retaining all
other LoRA updates. We compare the layer activations before and after
removing the module using the Fr\'echet Distance (FD), which measures both
the shift in the activation mean and the deformation in covariance
geometry~\citep{lucic2018gans}. The complete mathematical definition and
normalization are provided in Section~\ref{sec:methods}. For visualization,
we convert the resulting FD values into percentiles across all
layer-module pairs.

We further examine whether representation sensitivity is related to the
contribution of the adapter. Modules are ranked according to their
FD-based sensitivity, after which we progressively remove the
high-sensitivity, low-sensitivity, or randomly selected modules. We measure
the resulting increase in loss relative to the complete LoRA
adapter.

\begin{figure}[t]
    \raggedright
    \includegraphics[width=0.8\columnwidth]{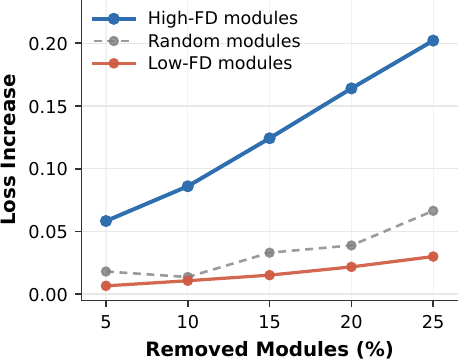}%
    \caption{Loss increase on BoolQ after removing different proportions of high-FD, random, or low-FD modules. Removing high-response modules consistently causes the largest functional degradation across different removal budgets.}
    \label{fig:boolq_removal}
\end{figure}


\paragraph{Observation 1: LoRA adapters exhibit strongly heterogeneous representation sensitivities across layers and modules.}
Figure~\ref{fig:boolq_fd_heatmap} shows that modules trained with the same
rank can affect the hidden representations to very different degrees.
Their sensitivity varies systematically across both layer depth and module
type. High-sensitivity modules are concentrated in particular
layer--module locations, while modules within the same layer can also show
large differences.

This result suggests that LoRA adaptation has a hierarchical structure.
Sensitivity varies across layers, and modules within each layer are not
equally sensitive. Adaptive rank allocation should therefore consider both
inter-layer and intra-layer differences rather than treating all module
locations uniformly.


\paragraph{Observation 2: Removing high-sensitivity modules causes greater
downstream task degradation.}
Figure~\ref{fig:boolq_removal} reports the validation-loss increase when
different groups of LoRA modules are progressively removed. Across all
removal ratios from \(5\%\) to \(25\%\), removing the high-sensitivity
modules causes substantially greater degradation than removing
low-sensitivity or randomly selected modules. At a removal ratio of $25\%$, the validation-loss increase for the high-sensitivity group exceeds $0.20$, which is approximately $2.9\times$ that of random removal ($0.07$) and $6.7\times$ that of low-sensitivity removal ($0.03$).

The separation also becomes larger as the removal ratio increases,
indicating that the result is not determined by only a few exceptional
modules. Instead, modules with higher representation sensitivity account
for a larger share of the function learned by the LoRA adapter. This
finding supports representation sensitivity as a useful signal for
adaptive rank allocation.

\subsection{From Post-training to Forward-only Probing}
\label{sec:motivation_transition}

The above observations show that representation sensitivity is both
structured and relevant to the downstream performance.
However, the sensitivity analyzed above is obtained from an already
trained LoRA adapter and
repeated module updates removal, making it unsuitable for rank
allocation before training. This leads to the central question of this work:

\textbf{\emph{Can representation sensitivity be estimated before fine-tuning, without significant computational cost?}}

RSRA addresses this question by estimating sensitivity before fine-tuning through standardized
virtual low-rank probing. By measuring the resulting
change in the hidden-state distribution, it obtains a forward-only
estimate of representation sensitivity and provides a task-aware rank allocation from a small task-specific
calibration set.

\section{Methodology: RSRA}
\label{sec:methods}

Let $\mathcal{L}$ denote the set of Transformer layers. For each layer
$l\in\mathcal{L}$, we adapt the linear modules
$\mathcal{J}_l=\{W_q,W_k,W_v,W_o,W_{\mathrm{gate}},
W_{\mathrm{up}},W_{\mathrm{down}}\}$.
RSRA uses a small task-specific calibration set to estimate a rank
configuration before fine-tuning. The entire allocation procedure relies
only on forward passes.

Motivated by the hierarchical sensitivity patterns observed in
Section~\ref{sec:motivation}, RSRA considers two complementary factors.
\emph{Inter-layer capacity} estimates how broadly representations are
distributed across spectral directions at each layer, while
\emph{intra-layer sensitivity} measures how strongly a module-specific
low-rank update changes the corresponding layer representations. These two
signals are combined to allocate a target average rank across all adapted
modules.

\subsection{Calibration Representations}
\label{sec:calibration_representation}



Let $X=\{x_i\}_{i=1}^{N}$ be a calibration set containing $N$
task-specific samples. For sample $x_i$, let $n_i$ denote the number
of valid tokens after excluding padding. After processing $X$ with
the base model, we collect the hidden-state vector
$h_{i,t}^{(l)}\in\mathbb{R}^{D}$ for each valid token $t$ at layer
$l$. We stack all valid-token representations into
$H^{(l)}=[h_{1,1}^{(l)},\ldots,h_{N,n_N}^{(l)}]^{\top}
\in\mathbb{R}^{T_l\times D}$, where
$T_l=\sum_{i=1}^{N}n_i$ is the total number of valid calibration tokens. Their empirical mean and covariance are computed as

\begin{equation}
\begin{aligned}
    \mu^{(l)}
    &=
    \frac{1}{T_l}
    \sum_{t=1}^{T_l} h_t^{(l)}, \\
    \Sigma^{(l)}
    &=
    \frac{1}{T_l-1}
    \sum_{t=1}^{T_l}
    \left(h_t^{(l)}-\mu^{(l)}\right)
    \left(h_t^{(l)}-\mu^{(l)}\right)^{\top}.
\end{aligned}
\end{equation}

\subsection{Inter-layer Allocation via Spectral Breadth}
\label{sec:inter_layer}

The preliminary analysis shows that representation sensitivity has a clear
depth-dependent structure. We therefore first assign a layer-level share of
the rank budget before distinguishing individual modules within each layer.

We introduce Spectral Effective Rank~\citep{roy2007erank} as a proxy for the
breadth of the representation space exposed by each layer. We first center the activation matrix:
\begin{equation}
    \widetilde{H}^{(l)}
    =
    H^{(l)}
    -
    \mathbf{1}
    \left(\mu^{(l)}\right)^{\top}.
\end{equation}

Let
\begin{equation}
    \widetilde{H}^{(l)}
    =
    U^{(l)}
    \operatorname{diag}
    \left(
        \sigma_1^{(l)},\ldots,\sigma_{q}^{(l)}
    \right)
    \left(V^{(l)}\right)^{\top}
\end{equation}
be its singular value decomposition, where
$q\leq\min(N,D)$. We normalize the singular values as
\begin{equation}
    p_i^{(l)}
    =
    \frac{\sigma_i^{(l)}}
    {\sum_{k=1}^{q}\sigma_k^{(l)}}.
\end{equation}

The effective rank of layer $l$ is defined as
\begin{equation}
    e_l
    =
    \operatorname{erank}
    \left(\widetilde{H}^{(l)}\right)
    =
    \exp
    \left(
        -\sum_{i=1}^{q}
        p_i^{(l)}\log p_i^{(l)}
    \right).
\end{equation}

A low effective rank indicates that the activation spectrum is concentrated
in a small number of directions, whereas a high effective rank indicates
that the representation is distributed across a broader spectral subspace.
RSRA therefore assigns a larger layer-level rank share to layers with
higher effective rank.

\subsection{Intra-layer Representation Sensitivity}
\label{sec:intra_layer}

We define \emph{representation sensitivity} as the change in a layer's
task-conditioned hidden-state distribution caused by a module-specific
low-rank update. To estimate this quantity before fine-tuning, RSRA
applies a standardized virtual low-rank update with probe rank $r_{\mathrm{probe}}$ to one
module at a time.


Consider a module
$W_{l,j}\in\mathbb{R}^{d_{\mathrm{out}}\times d_{\mathrm{in}}}$.
We independently sample two Gaussian factors
$A_{l,j}\in\mathbb{R}^{d_{\mathrm{out}}\times r_{\mathrm{probe}}}$ and
$B_{l,j}\in\mathbb{R}^{r_{\mathrm{probe}}\times d_{\mathrm{in}}}$, whose
entries are i.i.d.\ samples from $\mathcal{N}(0,1)$. Their product defines
a random probing direction with rank at most $r_{\mathrm{probe}}$. We
construct the virtual low-rank update as
\begin{equation}
    \Delta W_{l,j}
    =
    \epsilon
    \frac{\lVert W_{l,j}\rVert_F}
         {\lVert P_{l,j}\rVert_F}
    P_{l,j},
    \qquad
    P_{l,j}=A_{l,j}B_{l,j},
\end{equation}
where $\epsilon$ controls the update magnitude relative to the original
weight matrix. We set
$\epsilon=0.01$ throughout our experiments to approximate the relative
magnitude of a trained LoRA update while keeping the probe within a
small-perturbation regime.

For each module, we temporarily apply
$W_{l,j}\leftarrow W_{l,j}+\Delta W_{l,j}$ and perform a forward pass on
the same calibration set. Let
$H_{l,j}^{(l)\prime}$ denote the resulting output representation of layer
$l$, with empirical statistics
$\mu_{l,j}^{(l)\prime}$ and $\Sigma_{l,j}^{(l)\prime}$.

We compare the baseline and probed representation distributions using the
squared Fr\'echet Distance:
\begin{multline}
    d_{F}^2
    \left(
        (\mu^{(l)},\Sigma^{(l)}),
        (\mu_{l,j}^{(l)\prime},\Sigma_{l,j}^{(l)\prime})
    \right)
    =
    \|\mu^{(l)}-\mu_{l,j}^{(l)\prime}\|_2^2
    \\
    +
    \operatorname{Tr}
    \left(
        \Sigma^{(l)}
        +
        \Sigma_{l,j}^{(l)\prime}
        -
        2
        \left(
            (\Sigma^{(l)})^{\frac{1}{2}}
            \Sigma_{l,j}^{(l)\prime}
            (\Sigma^{(l)})^{\frac{1}{2}}
        \right)^{\frac{1}{2}}
    \right).
\end{multline}

The mean term measures the displacement of the activation center, while the
covariance term measures the change in the geometry and dispersion of the
representation distribution.

To make the sensitivity scores comparable across layers with different
activation scales, we normalize the Fr\'echet Distance by the second-order
energy of the baseline representation:
\begin{equation}
    s_{l,j}
    =
    \frac{
        d_F^2
        \left(
            (\mu^{(l)},\Sigma^{(l)}),
            (\mu_{l,j}^{(l)\prime},\Sigma_{l,j}^{(l)\prime})
        \right)
    }{
        \left\|\mu^{(l)}\right\|_2^2
        +
        \operatorname{Tr}
        \left(
            \Sigma^{(l)}
        \right)
    }.
\end{equation}
A larger $s_{l,j}$ means that the virtual low-rank update to module $j$
causes a larger change in the hidden-state distribution of layer $l$.

\subsection{Hierarchical Rank Allocation}
\label{sec:hierarchical_allocation}
RSRA combines the inter-layer spectral score and the intra-layer
sensitivity score through hierarchical normalization. The global
importance of module $j$ in layer $l$ is defined as
\begin{equation}
    I_{l,j}
    =
    \frac{
        e_l
    }{
        \sum_{k\in\mathcal{L}} e_k
    }
    \cdot
    \frac{
        s_{l,j}
    }{
        \sum_{m\in\mathcal{J}_l}s_{l,m}
    }.
\end{equation}
The first factor determines the share of the global rank budget assigned to
layer $l$, while the second factor distributes the layer-level budget among
its adapted modules.

Let
\begin{equation}
    \bar{I}
    =
    \frac{1}{M}
    \sum_{l\in\mathcal{L}}
    \sum_{j\in\mathcal{J}_l}
    I_{l,j},
    \qquad
    M
    =
    \sum_{l\in\mathcal{L}}
    |\mathcal{J}_l|,
\end{equation}
with a target average rank $\bar{r}$, we compute the rank assigned to each
module as
\begin{equation}
    R_{l,j}
    =
    \Pi_{[R_{\min},R_{\max}]}
    \left(
        \left\lfloor
        \frac{I_{l,j}}{\bar{I}}
        \cdot
        \bar{r}
        \right\rceil
    \right),
\end{equation}
where $\lfloor \cdot \rceil$ is the rounding operator to the nearest integer, and $\Pi_{\Omega}(\cdot)$ represents the projection operator onto the constraint set $\Omega$.

The resulting rank pattern prioritizes sensitive modules in layers with
broad representation spectra. Once the rank configuration is obtained,
the virtual updates are removed and standard LoRA fine-tuning proceeds
without any change to the optimization or inference procedure.

\section{Experiments}
\label{sec:experiment}

\begin{table*}[t] 
  \centering
  \small
  \setlength{\tabcolsep}{4pt} 
  \begin{tabular}{l l cccccccc c}
    \toprule
    \textbf{Model} & \textbf{Method} & \textbf{BoolQ} & \textbf{PIQA} & \textbf{SIQA} & \textbf{ARC-c} & \textbf{ARC-e} & \textbf{OBQA} & \textbf{HellaS} & \textbf{WinoG} & \textbf{Avg} \\
    \midrule
    \multirow{5}{*}{Qwen3-4B} 
    & LoRA    & 87.46{\scriptsize $\pm$0.07} & 79.16{\scriptsize $\pm$0.50} & 55.27{\scriptsize $\pm$0.44} & \underline{54.61}{\scriptsize $\pm$0.45} & \textbf{83.92}{\scriptsize $\pm$0.07} & \underline{45.20}{\scriptsize $\pm$0.37} & 56.77{\scriptsize $\pm$0.44} & 62.67{\scriptsize $\pm$0.50} & 65.63 \\
    & AdaLoRA & 84.65{\scriptsize $\pm$0.04} & 77.53{\scriptsize $\pm$0.41} & 52.30{\scriptsize $\pm$0.41} & 51.11{\scriptsize $\pm$1.31} & 82.53{\scriptsize $\pm$0.09} & 29.40{\scriptsize $\pm$0.50} & 53.13{\scriptsize $\pm$0.78} & \textbf{67.80}{\scriptsize $\pm$0.39} & 62.31 \\
    & SR-LoRA  & \underline{87.58}{\scriptsize $\pm$0.07} & \underline{79.82}{\scriptsize $\pm$0.41} & \underline{55.63}{\scriptsize $\pm$0.64} & \textbf{54.69}{\scriptsize $\pm$0.51} & 83.26{\scriptsize $\pm$0.07} & 45.00{\scriptsize $\pm$0.23} & \textbf{57.69}{\scriptsize $\pm$1.13} & 62.67{\scriptsize $\pm$0.30} & \underline{65.79} \\
    & GoRA    & 84.59{\scriptsize $\pm$0.15} & 69.48{\scriptsize $\pm$0.28} & 37.51{\scriptsize $\pm$1.06} & 48.81{\scriptsize $\pm$0.33} & 74.37{\scriptsize $\pm$0.11} & 26.60{\scriptsize $\pm$0.53} & 46.33{\scriptsize $\pm$0.98} & 52.09{\scriptsize $\pm$0.44} & 54.97 \\
    & \textbf{RSRA} & \textbf{87.74}{\scriptsize $\pm$0.08} & \textbf{79.87}{\scriptsize $\pm$0.24} & \textbf{55.73}{\scriptsize $\pm$0.67} & 54.44{\scriptsize $\pm$0.56} & \underline{83.63}{\scriptsize $\pm$0.12} & \textbf{45.80}{\scriptsize $\pm$0.40} & \underline{57.40}{\scriptsize $\pm$0.04} & \underline{63.22}{\scriptsize $\pm$0.24} & \textbf{65.98} \\
    \midrule
    \multirow{5}{*}{Mistral-7B} 
    & LoRA    & \underline{89.42}{\scriptsize $\pm$0.03} & \underline{84.22}{\scriptsize $\pm$1.16} & \underline{61.51}{\scriptsize $\pm$0.20} & 59.30{\scriptsize $\pm$0.32} & \underline{84.34}{\scriptsize $\pm$0.25} & \underline{52.00}{\scriptsize $\pm$0.31} & \textbf{69.39}{\scriptsize $\pm$0.73} & 59.75{\scriptsize $\pm$0.08} & \underline{69.99} \\
    & AdaLoRA & 85.81{\scriptsize $\pm$0.07} & 83.51{\scriptsize $\pm$0.51} & 59.88{\scriptsize $\pm$0.17} & 53.67{\scriptsize $\pm$0.63} & 80.68{\scriptsize $\pm$0.17} & 37.20{\scriptsize $\pm$0.27} & 65.30{\scriptsize $\pm$1.91} & \textbf{76.32}{\scriptsize $\pm$0.13} & 67.80 \\
    & SR-LoRA  & 89.69{\scriptsize $\pm$0.14} & 83.19{\scriptsize $\pm$1.39} & 60.54{\scriptsize $\pm$0.44} & \underline{59.56}{\scriptsize $\pm$0.16} & 84.51{\scriptsize $\pm$0.09} & \underline{52.00}{\scriptsize $\pm$0.59} & \underline{69.01}{\scriptsize $\pm$0.98} & 60.22{\scriptsize $\pm$0.09} & 69.84 \\
    & GoRA    & 47.49{\scriptsize $\pm$0.06} & 72.29{\scriptsize $\pm$0.01} & 34.19{\scriptsize $\pm$0.39} & 50.94{\scriptsize $\pm$0.24} & 80.22{\scriptsize $\pm$0.14} & 31.80{\scriptsize $\pm$0.27} & 55.43{\scriptsize $\pm$1.63} & 49.80{\scriptsize $\pm$0.15} & 52.77 \\
    & \textbf{RSRA} & \textbf{90.28}{\scriptsize $\pm$0.07} & \textbf{84.44}{\scriptsize $\pm$0.37} & \textbf{61.92}{\scriptsize $\pm$0.26} & \textbf{59.90}{\scriptsize $\pm$0.71} & \textbf{85.14}{\scriptsize $\pm$0.16} & \textbf{53.40}{\scriptsize $\pm$0.20} & 68.41{\scriptsize $\pm$1.12} & \underline{63.38}{\scriptsize $\pm$0.12} & \textbf{70.86} \\
    \bottomrule
  \end{tabular}
  \caption{Performance of Qwen3-4B and Mistral-7B with various rank allocation methods on commonsense reasoning benchmarks. Data are reported as mean $\pm$ std. The best performance within each LLM is indicated in bold, while the second best performance is highlighted in underline.}
  \label{tab:commonsense}
\end{table*}

We conduct comprehensive experiments to
evaluate the proposed \textbf{RSRA} through the following research questions:

\begin{itemize}
    \item \textbf{RQ1: Effectiveness.}
    Does representation-sensitive rank allocation improve parameter-efficient fine-tuning performance across different backbone models and downstream task domains?

    \item \textbf{RQ2:  Efficiency}
    What additional computation and memory cost does RSRA introduce, and how does it compare with existing rank-allocation methods?

    \item \textbf{RQ3: Compatibility and Extensibility.}
    Can RSRA serve as an effective plug-and-play rank allocator across diverse existing parameter-efficient fine-tuning variants?

    \item \textbf{RQ4: Validity of the Sensitivity Signal.}
    Does representation sensitivity estimated before fine-tuning predict the post-training contribution of adapted modules?
    
\end{itemize}


\subsection{Experimental Setup}
\label{sec:experimental_setup}

\paragraph{Benchmarks and Backbone Models.}
We evaluate RSRA on 16 datasets spanning two task families and
24 model--task settings. For
commonsense reasoning, Qwen3-4B~\citep{yang2025qwen3} and
Mistral-7B-v0.1~\citep{jiang2023mistral} are evaluated on BoolQ,
PIQA, SIQA, ARC-Easy, ARC-Challenge, OpenBookQA, HellaSwag, and
WinoGrande~\citep{clark2019boolq,bisk2020piqa,sap2019social,
clark2018arc,mihaylov2018obqa,zellers2019hellaswag,
sakaguchi2021winogrande}. For natural language understanding,
Mistral-7B-v0.1 is evaluated on the eight GLUE tasks~\citep{wang2018glue}.
We utilize the Language Model Evaluation Harness~\citep{gao2021lmeval} to conduct evaluations across commonsense reasoning and natural language understanding.

\paragraph{Baselines and Implementation Details.}
We compare RSRA with uniform LoRA and three representative
rank-allocation methods: AdaLoRA, SR-LoRA, and GoRA. To evaluate
plug-and-play compatibility, we further integrate RSRA with DoRA,
LoRA-FA, and PiSSA. For methods requiring a target or initial rank, we use a common setting $r=8$, with adapters applied to all attention and feed-forward linear
layers. Unless otherwise stated, results are averaged over three random
seeds. 
For efficiency, we report both standalone pre-allocation time and end-to-end
adaptation time. 
Since AdaLoRA adjusts ranks during fine-tuning, its allocation-related cost is approximated by the additional training time over vanilla LoRA.
Qwen3-4B experiments use four NVIDIA A10 24GB GPUs,
while Mistral-7B-v0.1 experiments use one NVIDIA A800 80GB GPU.

\subsection{RQ1: Effectiveness across Models and Tasks}
\label{sec:rq1_effectiveness}

We first investigate whether RSRA achieves competitive and consistent
downstream performance across different backbone models and task
domains. Tables~\ref{tab:commonsense} and~\ref{tab:nlu} compare RSRA
with LoRA and representative adaptive rank-allocation methods
on commonsense reasoning and natural language understanding benchmarks.

\paragraph{RSRA delivers the strongest aggregate performance across
the evaluated settings.}
On commonsense reasoning, RSRA obtains average accuracies of 65.98
and 70.86 with Qwen3-4B and Mistral-7B, respectively, exceeding the
strongest competing results by 0.19 and 0.87 points. On GLUE, RSRA
achieves an average score of 77.58, outperforming SR-LoRA and uniform
LoRA by 1.20 and 2.88 points, respectively. Notably, RSRA is the only
evaluated method that achieves the highest average performance in all
three model--benchmark settings, demonstrating strong and consistent
effectiveness across backbone models and task domains.

\paragraph{RSRA provides broad task-level gains while revealing the
task relevance of rank allocation.}
Across the two commonsense backbones, RSRA achieves the best result
on 10 of 16 model--task combinations and obtains the best or tied-best
result on 6 of 8 GLUE tasks. Overall, it ranks first on 16 of the 24 evaluated settings, showing that its aggregate
advantage is supported by broad improvements across heterogeneous
tasks rather than isolated gains. Meanwhile, the stronger performance
of AdaLoRA on WinoGrande and GoRA on RTE and WNLI indicates that
different tasks may favor distinct allocation patterns. This observation
further highlights the task-relevant nature of rank allocation and
the advantage of RSRA in providing a consistently strong allocation
strategy across diverse adaptation settings.

\begin{table*}[t]
  \centering
  \small 
  
  \setlength{\tabcolsep}{5pt} 
  \begin{tabular}{l cccccccc c}
    \toprule
    \textbf{Method} & \textbf{CoLA} & \textbf{RTE} & \textbf{SST2} & \textbf{WNLI} & \textbf{MNLI} & \textbf{MRPC} & \textbf{QNLI} & \textbf{QQP} & \textbf{Avg} \\
    \midrule
    LoRA    & \underline{68.47}{\scriptsize $\pm$0.10} & 52.71{\scriptsize $\pm$0.16} & \textbf{96.56}{\scriptsize $\pm$0.09} & 43.66{\scriptsize $\pm$0.12} & \underline{88.85}{\scriptsize $\pm$0.16} & 68.38{\scriptsize $\pm$0.04} & \underline{91.23}{\scriptsize $\pm$0.14} & 87.75{\scriptsize $\pm$0.20} & 74.70 \\
    AdaLoRA & 58.33{\scriptsize $\pm$0.06} & \underline{66.43}{\scriptsize $\pm$0.07} & \textbf{96.56}{\scriptsize $\pm$0.11} & \underline{56.34}{\scriptsize $\pm$0.08} & 85.76{\scriptsize $\pm$0.14} & 73.53{\scriptsize $\pm$0.11} & 84.73{\scriptsize $\pm$0.06} & 85.16{\scriptsize $\pm$0.24} & 75.86 \\
    SR-LoRA  & 67.69{\scriptsize $\pm$0.08} & 52.71{\scriptsize $\pm$0.14} & \underline{96.44}{\scriptsize $\pm$0.05} & 43.66{\scriptsize $\pm$0.14} & 88.69{\scriptsize $\pm$0.12} & \underline{82.84}{\scriptsize $\pm$0.06} & 91.10{\scriptsize $\pm$0.01} & \underline{87.88}{\scriptsize $\pm$0.18} & \underline{76.38} \\
    GoRA    & 54.00{\scriptsize $\pm$0.10} & \textbf{76.53}{\scriptsize $\pm$0.03} & 90.71{\scriptsize $\pm$0.12} & \textbf{69.01}{\scriptsize $\pm$0.09} & 57.69{\scriptsize $\pm$0.15} & 75.61{\scriptsize $\pm$0.18} & 54.46{\scriptsize $\pm$0.13} & 76.26{\scriptsize $\pm$0.13} & 69.28 \\
    \textbf{RSRA} & \textbf{69.98}{\scriptsize $\pm$0.13} & 52.71{\scriptsize $\pm$0.05} & \textbf{96.56}{\scriptsize $\pm$0.04} & 44.72{\scriptsize $\pm$0.08} & \textbf{89.09}{\scriptsize $\pm$0.17} & \textbf{84.56}{\scriptsize $\pm$0.09} & \textbf{91.47}{\scriptsize $\pm$0.02} & \textbf{91.51}{\scriptsize $\pm$0.19} & \textbf{77.58} \\
    \bottomrule
   \end{tabular}
   \caption{Performance of Mistral-7B with various rank allocation methods on the GLUE benchmarks, reported as mean $\pm$ std.}
   \label{tab:nlu}
\end{table*}

\subsection{RQ2: Efficiency and Practicality}
\label{sec:rq2_efficiency}

\begin{figure}[t]
    \centering
\includegraphics[width=0.8\columnwidth]{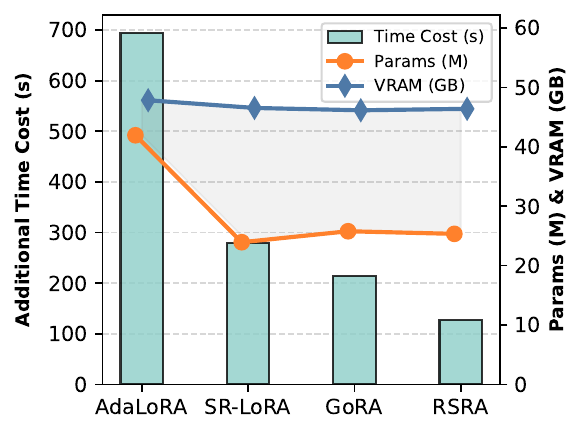}
    \centerline{(a) PIQA}
\includegraphics[width=0.8\columnwidth]{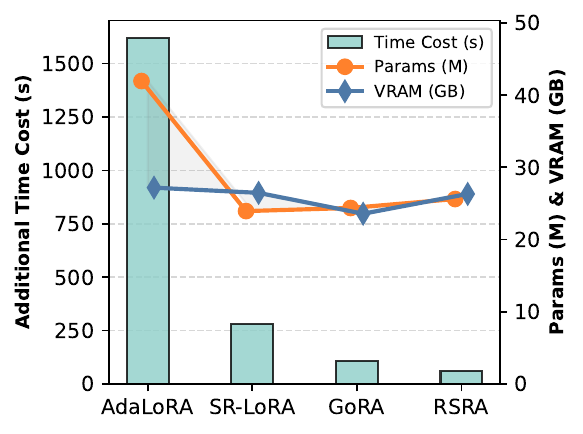}
    \centerline{(b) CoLA}
    \caption{
    Efficiency comparison on (a) PIQA and (b) CoLA.
    Bars report the rank-allocation additional time, while the orange
    and blue lines denote the number of trainable parameters and peak
    GPU memory, respectively.
    }
    \label{fig:efficiency}
\end{figure}

We next investigate the computational efficiency of RSRA and the
additional cost introduced by task-aware rank allocation.
Table~\ref{tab:efficiency} reports the standalone pre-allocation time,
end-to-end adaptation time, and relative overhead over LoRA
across different backbone models and task domains.
Figure~\ref{fig:efficiency} further compares the wall-clock cost
associated with rank allocation, trainable parameter count, and peak
GPU memory on representative commonsense reasoning and natural
language understanding tasks. 

\begin{table}[t]
  \centering
  \small
  \setlength{\tabcolsep}{2.8mm} 

  \begin{tabular}{lccc}
    \toprule
    \textbf{Method} & \textbf{Alloc. Time} & \textbf{E2E Time} & \textbf{E2E Overhead} \\
    \midrule

    \rowcolor{gray!20}
    \multicolumn{4}{c}{\textbf{Commonsense Reasoning (Qwen3-4B)}} \\ 
    LoRA     & --      & 17m24s & -- \\
    AdaLoRA  & --      & 26m45s & +53.7\% \\
    SR-LoRA  & 169.38s & 21m16s & +22.2\% \\
    GoRA     & 122.00s & 20m28s & +17.6\% \\
    \textbf{RSRA} & \textbf{82.40s} & \textbf{19m31s} & \textbf{+12.2\%} \\
    \midrule

    \rowcolor{gray!20}
    \multicolumn{4}{c}{\textbf{Commonsense Reasoning (Mistral-7B)}} \\ 
    LoRA     & --      & 27m36s & -- \\
    AdaLoRA  & --      & 38m24s & +39.1\% \\
    SR-LoRA  & 278.70s & 33m11s & +20.2\% \\
    GoRA     & 241.98s & 33m17s & +20.6\% \\
    \textbf{RSRA} & \textbf{127.90s} & \textbf{30m08s} & \textbf{+9.2\%} \\
    \midrule

    \rowcolor{gray!20}
    \multicolumn{4}{c}{\textbf{Natural Language Understanding (Mistral-7B)}} \\ 
    LoRA     & --      & 22m29s & -- \\
    AdaLoRA  & --      & 30m28s & +35.5\% \\
    SR-LoRA  & 278.70s & 27m15s & +21.2\% \\
    GoRA     & 238.36s & 26m54s & +19.6\% \\
    \textbf{RSRA} & \textbf{123.20s} & \textbf{24m08s} & \textbf{+7.3\%} \\
    \bottomrule
  \end{tabular}

  \caption{Wall-clock efficiency of different rank allocation methods. Alloc. Time denotes the standalone pre-allocation time, while E2E Time includes both rank allocation and fine-tuning. E2E Overhead is measured relative increase to LoRA.}
  \label{tab:efficiency}
\end{table}

\paragraph{Forward-only probing enables substantially faster rank
allocation.}
RSRA achieves the lowest standalone allocation time across all three
evaluated settings, requiring only 82.40 seconds on Qwen3-4B and
127.90 and 123.20 seconds on Mistral-7B for commonsense reasoning and
natural language understanding, respectively. Compared with the fastest
competing pre-allocation method, RSRA provides allocation speedups of
1.48$\times$, 1.89$\times$, and 1.93$\times$ across the three settings.
This advantage consistently translates into the shortest end-to-end
time among all adaptive rank-allocation methods: 19m31s on Qwen3-4B,
30m08s on Mistral-7B commonsense reasoning, and 24m08s on Mistral-7B
natural language understanding. These results demonstrate that
forward-only representational probing can identify task-relevant rank
configurations considerably more efficiently than other allocation strategies.

\paragraph{RSRA reduces allocation cost without inflating adapter capacity.}
As shown in Figure~\ref{fig:efficiency}, RSRA maintains a trainable
parameter budget comparable to SR-LoRA and GoRA, while its peak GPU
memory remains within a similar range across both PIQA and CoLA.
Notably, RSRA does not consistently use fewer trainable parameters
than the competing pre-allocation methods, yet still requires the
lowest rank-allocation cost.
The observed wall-clock improvement therefore stems from the efficiency
of its forward-only probing procedure, rather than from a materially
smaller adapter budget or a reduction in memory consumption.

\definecolor{lightgray}{gray}{0.9}
\begin{table}[t]
\centering
\setlength{\tabcolsep}{1.2mm} 

{\small
\begin{tabular}{lcccccc}
\toprule
\textbf{Method}
& \textbf{MNLI}
& \textbf{MRPC}
& \textbf{QNLI}
& \textbf{QQP}
& \textbf{ARC-C}
& \textbf{OBQA} \\
\midrule

DoRA
& 88.95
& 80.39
& 90.96
& \textbf{87.67}
& 59.15
& 52.60 \\

\rowcolor{lightgray}
+RSRA
& \textbf{89.04}
& \textbf{83.33}
& \textbf{91.20}
& 87.61
& \textbf{60.90}
& \textbf{53.40} \\

\addlinespace[0.25em]

LoRA-FA
& \textbf{88.76}
& 81.62
& 90.12
& \textbf{87.28}
& 58.36
& 50.00 \\

\rowcolor{lightgray}
+RSRA
& 88.45
& \textbf{83.33}
& \textbf{90.59}
& 87.26
& \textbf{59.39}
& \textbf{51.40} \\

\addlinespace[0.25em]

PiSSA
& 88.55
& 81.22
& 48.38
& 40.33
& 22.95
& 51.20 \\

\rowcolor{lightgray}
+RSRA
& \textbf{89.12}
& \textbf{82.11}
& \textbf{90.77}
& \textbf{87.53}
& \textbf{57.59}
& \textbf{54.40} \\

\bottomrule
\end{tabular}
}

\caption{Compatibility and extensibility of RSRA across representative PEFT variants. Shaded rows denote each PEFT method after integrating RSRA rank allocation. 
}
\label{tab:integratability}
\end{table}

\subsection{RQ3: Compatibility and Extensibility}
\label{sec:rq4_integratability}

Since RSRA operates at the rank-allocation level without redesigning
the underlying adapter, we next investigate whether it can serve as an
effective plug-and-play allocator across diverse PEFT methods.
To this end, we integrate RSRA into DoRA, LoRA-FA, and PiSSA, which
respectively modify adapter parameterization, optimization, and
initialization, while retaining their original adaptation mechanisms.
Table~\ref{tab:integratability} reports the resulting performance across
natural language understanding and commonsense reasoning benchmarks.


\paragraph{RSRA serves as an effective plug-and-play rank allocator across
diverse PEFT designs.}
RSRA can be directly integrated into DoRA, LoRA-FA, and PiSSA
without altering their original parameterization, optimization, or
initialization mechanisms. Across the 18 evaluated task--variant
combinations, integrating RSRA improves 15 settings and increases
the average performance of all three PEFT methods. Specifically, the
average scores of DoRA and LoRA-FA increase from 76.62 to 77.58 and
from 76.02 to 76.74, respectively. These consistent gains across both
natural language understanding and commonsense reasoning demonstrate
that RSRA is not tied to a particular PEFT design and can be readily
deployed as a plug-and-play allocation module.

\paragraph{RSRA preserves strong configurations while substantially
improving poorly matched ones.}
For the already competitive DoRA and LoRA-FA configurations, RSRA
largely preserves their strong results while providing consistent gains
on MRPC, QNLI, ARC-C, and OBQA. The improvement is more pronounced
for PiSSA, whose average score increases from 55.44 to 76.92.
In particular, RSRA improves QNLI, QQP, and ARC-C by 42.39, 47.20,
and 34.64 points, respectively, while also improving the remaining
three tasks. This pattern indicates that task-aware rank allocation
can complement strong PEFT configurations and effectively correct rank
distributions that are poorly matched to downstream adaptation demands,
supporting RSRA as a broadly extensible allocation module.

\subsection{RQ4: Validity of the Sensitivity Signal}
\label{sec:rq4_validity}

We finally examine whether the representation-sensitivity signal estimated
before fine-tuning is predictive of module contribution after adaptation.
For each task, we rank all adapted modules according to their pre-fine-tuning
RSRA sensitivity scores and select the top, middle, and bottom 10\% groups.
After fine-tuning, we independently ablate each group by removing its trained
low-rank updates while keeping all remaining adapter updates fixed.
We quantify the contribution of an ablated group using the increase in
test loss,
\begin{equation}
    \Delta\mathcal{L}
    =
    \mathcal{L}_{\mathrm{ablated}}
    -
    \mathcal{L}_{\mathrm{full}},
\end{equation}
where a larger $\Delta\mathcal{L}$ indicates that the removed modules make
a greater contribution to the adapted model.

\begin{figure}[t]
    \centering
    \includegraphics[width=0.95\columnwidth]{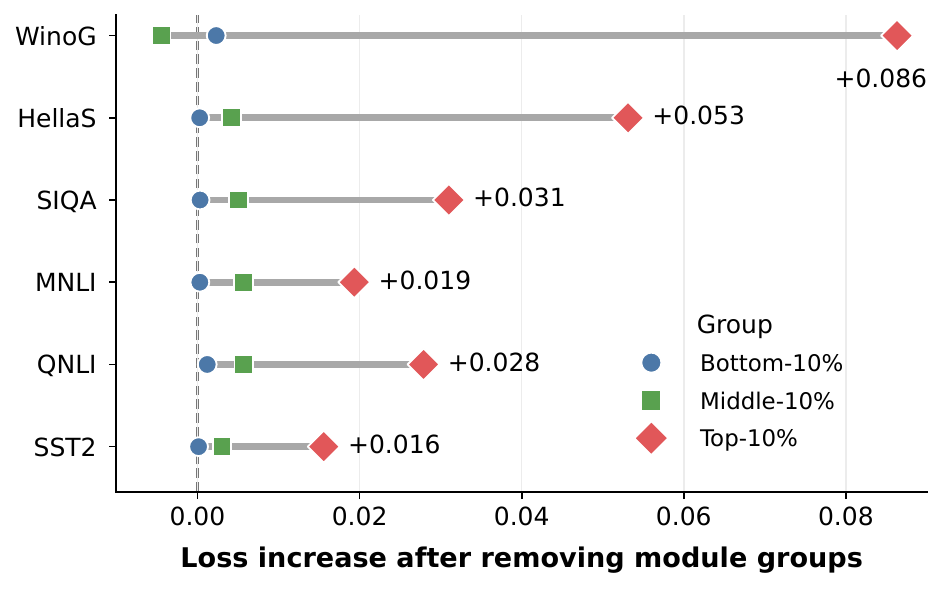}
    \caption{
    Validation of the representation-sensitivity scores
    across three commonsense reasoning and three GLUE tasks; annotations denote the top--bottom gap.
    Removing the trained updates of the top-ranked 10\% of modules causes
    the largest loss increase on every task.
    }
    \label{fig:group_ablation}
\end{figure}

\paragraph{RSRA sensitivity predicts post-training module contribution.}
As shown in Figure~\ref{fig:group_ablation}, ablating the top-ranked 10\%
of modules produces the largest loss increase on all six tasks,
consistently exceeding the effects of ablating the middle- or bottom-ranked
groups. Compared with the bottom-ranked group, the top-ranked group yields
loss-increase gaps of 0.086 on WinoGrande, 0.053 on HellaSwag, 0.031 on
SIQA, 0.019 on MNLI, 0.028 on QNLI, and 0.016 on SST-2. Because the module
ranking is determined before fine-tuning, whereas module contribution is
measured only after adaptation, this consistent ordering provides independent
evidence that the forward-only sensitivity estimates capture differences
that remain relevant to the trained adapter.

\paragraph{The sensitivity--contribution alignment holds across task families.}
The same pattern is observed on both commonsense reasoning and natural
language understanding benchmarks, despite their different task objectives
and data characteristics. Although the magnitude of the top--bottom gap
varies across tasks, the most sensitive group is consistently the most
consequential after fine-tuning. These results suggest that RSRA captures
a task-relevant allocation signal rather than a pattern specific to a
single benchmark or task family, supporting its use for distributing
adaptation capacity before optimization.

\section{Conclusion}
\label{sec:conclusion}

In this paper, we reveal structured representation sensitivity in LoRA
adaptation and propose RSRA, a training-free forward-only rank allocator
via representation probing. Experiments demonstrate that RSRA enables
effective and efficient rank allocation with broad compatibility across
diverse models and PEFT methods.

\bibliography{aaai2027}

\clearpage
\appendix

\section{Related Work}
\subsection{LoRA and LoRA Variants}

Parameter-efficient fine-tuning (PEFT) aims to adapt large
pre-trained models by updating only a small subset of parameters.
Among existing PEFT approaches, Low-Rank Adaptation (LoRA)
\citep{hu2022lora} has become one of the most widely adopted methods
for large language models. LoRA represents parameter updates as
low-rank matrix decompositions, substantially reducing trainable
parameters and computational cost while maintaining competitive
adaptation performance.

Following LoRA, numerous variants have been proposed to improve
adaptation effectiveness and efficiency. AdaLoRA\citep{zhang2023adalora} dynamically adjusts rank allocation during
fine-tuning through singular value decomposition. 
HiRA\citep{huang2025hira} introduces Hadamard High-Rank Adaptation to alleviate the
performance saturation of conventional LoRA when increasing the rank.
DoRA\citep{liu2024dora} improves adaptation by decomposing LoRA updates
into magnitude and directional components. LoRA-FA\citep{zhang2023lora_fa} reduces memory overhead by freezing one
factor of the low-rank update. 
PiSSA\citep{meng2024pissa} initializes LoRA updates using principal
singular vectors of pretrained weights to improve optimization.

Despite these improvements, the rank configuration remains a critical
factor affecting LoRA performance. Uniform rank assignment ignores the
heterogeneity among Transformer layers and modules, motivating
adaptive rank allocation methods.

\subsection{Adaptive Rank Allocation for LoRA}

Adaptive rank allocation methods aim to distribute a fixed parameter
budget across different LoRA modules according to their importance.
Existing approaches can be categorized into training-based allocation
and pre-allocation methods.

\paragraph{Training-based Allocation.}

Training-based approaches determine rank allocation dynamically during
the fine-tuning process by leveraging optimization signals.
AdaLoRA
\citep{zhang2023adalora} is an early representative method that
parameterizes LoRA updates with singular value decomposition and
dynamically prunes less important singular components during training.
DyLoRA
\citep{valipour2022dylora} enables flexible rank adaptation by
training LoRA adapters with randomly sampled ranks.
SoRA
\citep{ding2023sora} introduces sparse optimization with learnable
gates to identify unnecessary low-rank components.
ALoRA
\citep{liu2024alora} further explores adaptive allocation through
module importance estimation and parameter budget redistribution.
AutoLoRA
\citep{zhang2024autolora}
formulates rank selection as a meta-learning problem and automatically
tunes matrix ranks according to downstream feedback.

These approaches provide flexible rank adjustment but generally require
additional optimization procedures, which may introduce extra training
overhead.

\paragraph{Pre-allocation Methods.}

Pre-allocation methods determine rank configurations before fine-tuning,
avoiding iterative rank adjustment during optimization.
SR-LoRA
\citep{zhang2025srlora} allocates ranks according to the stable rank of
pretrained weight matrices, using intrinsic weight statistics as a proxy
for adaptation capacity.
AIRA
\citep{li2025aira} incorporates activation information by analyzing
activation distributions from calibration data and assigning ranks
according to activation characteristics.
GoRA
\citep{he2025gora} estimates module importance using gradient signals
collected from a small calibration set and determines rank allocation
before fine-tuning.

Compared with training-based methods, pre-allocation approaches avoid
dynamic rank search during adaptation and provide better scalability
for large language models. However, their allocation criteria rely on
different proxies, including weight statistics, activation patterns,
or gradient-based importance.

\begin{table*}[t]
  \centering
  \setlength{\tabcolsep}{4pt} 
  \begin{tabular}{ll cccccccc}
    \toprule
    \textbf{Model} & \textbf{Method} & \textbf{Rank} & \textbf{LR} & \textbf{Batchsize} & \textbf{Epochs} & \textbf{Max len} & \textbf{Warmup} & \textbf{Scheduler} & \textbf{Optim.} \\

    \rowcolor{gray!10} \multicolumn{10}{c}{\textit{Task: Commonsense Reasoning}} \\
    \multirow{4}{*}{Qwen3-4B} 
    & LoRA    & \multirow{4}{*}{8} & \multirow{4}{*}{2e-4} & \multirow{4}{*}{64} & \multirow{4}{*}{3} & \multirow{4}{*}{512} & \multirow{4}{*}{0.1} & \multirow{4}{*}{Cosine} & \multirow{4}{*}{AdamW} \\
    & AdaLoRA & & & & & & & & \\
    & SR-LoRA & & & & & & & & \\
    & GoRA    & & & & & & & & \\
    \cmidrule{1-2}
    \multirow{4}{*}{Mistral-7B} 
    & LoRA    & \multirow{4}{*}{8} & \multirow{4}{*}{2e-4} & \multirow{4}{*}{128} & \multirow{4}{*}{3} & \multirow{4}{*}{512} & \multirow{4}{*}{0.1} & \multirow{4}{*}{Cosine} & \multirow{4}{*}{AdamW} \\
    & AdaLoRA & & & & & & & & \\
    & SR-LoRA & & & & & & & & \\
    & GoRA    & & & & & & & & \\
    \midrule

    \rowcolor{gray!10} \multicolumn{10}{c}{\textit{Task: Mathematical Reasoning}} \\
    \multirow{4}{*}{\makecell[l]{Deepseek-\\R1-1.5B} }
    & LoRA    & \multirow{4}{*}{8} & \multirow{4}{*}{2e-4} & \multirow{4}{*}{64} & \multirow{4}{*}{4} & \multirow{4}{*}{1024} & \multirow{4}{*}{0.1} & \multirow{4}{*}{Cosine} & \multirow{4}{*}{AdamW} \\
    & AdaLoRA & & & & & & & & \\
    & SR-LoRA & & & & & & & & \\
    & GoRA    & & & & & & & & \\
    \midrule
    \rowcolor{gray!10} \multicolumn{10}{c}{\textit{Task: Natural Language Understanding}} \\
    \multirow{4}{*}{Mistral-7B} 
    & LoRA    & \multirow{4}{*}{8} & \multirow{4}{*}{2e-4} & \multirow{4}{*}{128} & \multirow{4}{*}{3} & \multirow{4}{*}{256} & \multirow{4}{*}{0.1} & \multirow{4}{*}{Cosine} & \multirow{4}{*}{AdamW} \\
    & AdaLoRA & & & & & & & & \\
    & SR-LoRA & & & & & & & & \\
    & GoRA    & & & & & & & & \\
    \midrule
    \bottomrule
  \end{tabular}
    \caption{Hyperparameter settings for different models and tasks across all baselines.}
  \label{tab:all_baselines}
\end{table*}

\section{Experiment Details}
\subsection{Experimental Settings}
\paragraph{Commonsense Reasoning} We evaluate our method on eight representative benchmarks: BoolQ\citep{clark2019boolq}, PIQA\citep{bisk2020piqa}, SIQA\citep{sap2019social}, OBQA\citep{mihaylov2018obqa}, HellaSwag\citep{zellers2019hellaswag}, WinoGrande\citep{sakaguchi2021winogrande}, ARC-e and ARC-c\citep{clark2018arc}. Qwen3-4B\citep{yang2025qwen3} and Mistral-7B-v0.1\citep{jiang2023mistral} are utilized as the base models. Experiments for Qwen3-4B were conducted on a node with four NVIDIA A10 24GB GPUs, while Mistral-7B-v0.1 was trained and evaluated on a single NVIDIA A800 80GB GPU. All evaluations are performed using the Language Model Evaluation Harness~\cite{gao2021lmeval} to ensure standardized zero-shot assessment.

\paragraph{General Language Understanding} We evaluate our method on the GLUE benchmark~\cite{wang2018glue} across eight sub-tasks: MNLI, QNLI, QQP, RTE, SST-2, MRPC, CoLA, and WNLI. We employ Mistral-7B-v0.1 as the backbone model for these natural language understanding tasks. All training and evaluation processes were conducted on a single NVIDIA A100 80GB GPU. For all sub-tasks, we report the classification accuracy as the evaluation metric to provide a consistent measure of performance across the benchmark.

\paragraph{Mathematical Reasoning} We conduct experiments on the GSM8K~\cite{cobbe2021gsm8k} dataset and five sub-tasks from the MATH (Hendrycks) benchmark~\cite{hendrycks2021}, specifically: Algebra, Geometry, Precalculus, Counting \& Probability, and Number Theory. For these tasks, we employ the DeepSeek-R1-Distill-Qwen-1.5B model\citep{guo2025deepseek}. Training and evaluation were performed on a cluster of four NVIDIA A10 24GB GPUs. For evaluation, we utilize the Language Model Evaluation Harness~\cite{gao2021lmeval} for GSM8K, while the procedure for the MATH benchmark follows the official settings and configurations as established in~\cite{hendrycks2021}.

\subsection{Hyperparameter Analysis and Reproducibility}
\label{sec:appendix_hyperparams}

To ensure a fair comparison and reproducibility, we provide a detailed breakdown of the hyperparameter configurations used in our experiments. 

\begin{table}[t]
  \centering
  \small
  \begin{tabular}{cccccc}
    \toprule
    \textbf{Average Rank} & \textbf{$r_{probe}$} & \textbf{$\epsilon$} & \textbf{$N$} & \textbf{$R_{min}$} & \textbf{$R_{max}$} \\
    \midrule
    8 & 8 & 0.01 & 256 & 2 & 128 \\
    \bottomrule
  \end{tabular}
  \caption{Specific hyperparameters for the proposed RSRA method.}
  \label{tab:RSRA_hyperparams}
\end{table}

\paragraph{General Training Configurations} 
As summarized in Table~\ref{tab:all_baselines}, we maintain consistent training hyperparameters across all methods (LoRA, AdaLoRA, SR-LoRA, GoRA, and our RSRA) for each specific task and model. We use the AdamW optimizer with a cosine learning rate scheduler. The peak learning rate is set to $2 \times 10^{-4}$ for most tasks, which has been empirically found to be stable for PEFT. For Commonsense Reasoning and NLU tasks, we train for 3 epochs, while for Mathematical Reasoning, we extend the training to 4 epochs to allow for better convergence on complex logic. The batch size is adjusted (64 or 128) based on the GPU memory capacity of the respective hardware mentioned in Section Experimental Settings.

\begin{table*}[t]
  \centering
  \setlength{\tabcolsep}{6pt} 
  \begin{tabular}{l cccccc c}
    \toprule
    \textbf{Method} & \textbf{GSM8K} & \textbf{Algebra} & \textbf{C\&P} & \textbf{Geometry} & \textbf{Precalc} & \textbf{NumThy} & \textbf{Avg} \\
    \midrule
    LoRA    & 60.42{\scriptsize $\pm$0.24} & \underline{53.03}{\scriptsize $\pm$0.36} & 30.59{\scriptsize $\pm$1.06} & \underline{27.56}{\scriptsize $\pm$0.17} & 16.12{\scriptsize $\pm$0.94} & 34.07{\scriptsize $\pm$1.45} & 36.97 \\
    AdaLoRA & \underline{61.41}{\scriptsize $\pm$0.56} & 51.47{\scriptsize $\pm$0.55} & \underline{32.28}{\scriptsize $\pm$0.69} & 25.68{\scriptsize $\pm$0.45} & \textbf{19.23}{\scriptsize $\pm$1.19} & \textbf{50.19}{\scriptsize $\pm$1.24} & \textbf{40.04} \\
    SR-LoRA & 60.35{\scriptsize $\pm$0.29} & 52.99{\scriptsize $\pm$0.13} & 31.43{\scriptsize $\pm$0.76} & \underline{27.56}{\scriptsize $\pm$0.57} & 16.12{\scriptsize $\pm$1.24} & 31.48{\scriptsize $\pm$2.13} & 36.66 \\
    GoRA    & 50.80{\scriptsize $\pm$0.76} & 49.37{\scriptsize $\pm$0.41} & 21.10{\scriptsize $\pm$0.56} & 17.95{\scriptsize $\pm$0.39} & 9.16{\scriptsize $\pm$0.75} & 29.07{\scriptsize $\pm$1.14} & 29.58 \\
    \textbf{RSRA} & \textbf{62.55}{\scriptsize $\pm$0.32} & \textbf{53.37}{\scriptsize $\pm$0.54} & \textbf{35.44}{\scriptsize $\pm$0.47} & \textbf{27.77}{\scriptsize $\pm$0.99} & \underline{16.67}{\scriptsize $\pm$0.95} & \underline{34.81}{\scriptsize $\pm$1.65} & \underline{38.44} \\
    \bottomrule
    \multicolumn{8}{l}{\footnotesize \textit{Note:} C\&P: Counting \& Probability; Precalc: Precalculus; NumThy: Number Theory.} \\
  \end{tabular}
  \caption{Performance of DeepSeek-R1 with various rank allocation methods on mathematical reasoning benchmarks. Data are reported as mean $\pm$ std.}
  \label{tab:math}
\end{table*}

\paragraph{RSRA Specific Parameters}
Table~\ref{tab:RSRA_hyperparams} lists the specific hyperparameters introduced by our RSRA framework. These parameters govern the training-free rank discovery process:
\begin{itemize}
    \item \textbf{Probe Rank ($r_p$):} We set the virtual probing rank to 8. This rank is used to calculate the representational sensitivity and does not involve any actual parameter updates.
    \item \textbf{Calibration Samples ($N$):} A small set of $N=256$ samples is used to compute the sensitivity scores. Our experiments indicate that RSRA is robust to the choice of $N$, and 256 samples are sufficient to capture the activation-space geometry.
    \item \textbf{Rank Boundaries ($R_{min}, R_{max}$):} To prevent extreme rank allocation (such as assigning zero rank to a layer or exceeding hardware memory limits), we constrain the discovered rank within the range $[2, 128]$. The average rank across all layers is maintained at $r=8$ to ensure a fair parameter budget comparison with the LoRA baseline.
\end{itemize}
The stability of these parameters suggests that \textbf{RSRA} does not require extensive per-task tuning, making it a "plug-and-play" solution for automatic rank allocation.

\section{Additional Experiments}
\subsection{Mathematical Reasoning}

To further evaluate the generalizability of RSRA beyond commonsense
reasoning and natural language understanding tasks, we conduct additional
experiments on mathematical reasoning benchmarks using DeepSeek-R1.
We compare RSRA with representative rank allocation baselines, including
LoRA, AdaLoRA, SR-LoRA, and GoRA, across six mathematical reasoning
categories: GSM8K, Algebra, Counting \& Probability, Geometry,
Precalculus, and Number Theory. All results are averaged over multiple
random seeds and reported as mean $\pm$ standard deviation.

As shown in Table~\ref{tab:math}, RSRA achieves competitive performance
across diverse mathematical reasoning tasks and obtains the best results
on GSM8K, Algebra, Counting \& Probability, and Geometry. Compared with
uniform LoRA, RSRA improves the average performance from 36.97 to 38.44,
demonstrating that representation-sensitive rank allocation can provide
effective adaptation capacity for reasoning-oriented tasks. Although
AdaLoRA achieves higher average performance by benefiting from dynamic
rank adjustment during fine-tuning, RSRA maintains strong performance
with a training-free allocation procedure, outperforming other
pre-allocation baselines including SR-LoRA and GoRA.

\subsection{Detailed Efficiency Analysis}
\label{sec:appendix_efficiency}

\begin{table}[htbp]
  \centering
  \small
  \setlength{\tabcolsep}{3.5pt} 
  \begin{tabular}{l ccccc}
    \toprule
    Method & Add. Time & Peak VRAM & Params & Acc. & $\Delta$ Acc. \\
           & (seconds) & (MB)      & (M)    & (\%) & (\%) \\
    \midrule
    
    \rowcolor{gray!15} 
    \multicolumn{6}{c}{\textbf{PIQA}} \\ 
    
    LoRA & - & 47,815 & 41.94 & 84.22 & - \\
    AdaLoRA & 695.3 & 48,995 & 41.95 & 83.51 & \textcolor{blue}{-0.71} \\
    SR-LoRA & 278.7 & 47,667 & \textbf{23.94} & 83.19 & \textcolor{blue}{-1.03} \\
    GoRA & 215.3 & \textbf{47,263} & 25.7 & 72.29 & \textcolor{blue}{-11.93} \\
    \textbf{RSRA} & \textbf{128.0} & 47,517 & 25.35 & \textbf{84.44} & \textcolor{red}{+0.22} \\
    
    \midrule
    
    \rowcolor{gray!15} 
    \multicolumn{6}{c}{\textbf{CoLA}} \\
    
    LoRA & - & 27,029 & 41.94 & 68.47 & - \\
    AdaLoRA & 1620.8 & 27,837 & 41.95 & 58.33 & \textcolor{blue}{-10.14} \\
    SR-LoRA & 278.7 & 27,101 & \textbf{23.94} & 67.69 & \textcolor{blue}{-0.78} \\
    GoRA & 106.6 & \textbf{24,140} & 24.37 & 54.00 & \textcolor{blue}{-14.47} \\
    \textbf{RSRA} & \textbf{62.3} & 26,947 & 25.61 & \textbf{69.98} & \textcolor{red}{+1.51} \\
    \bottomrule
  \end{tabular}
  \caption{Efficiency and performance comparison of various rank allocation methods on Mistral-7B. Add. Time denotes the additional training time overhead incurred by the allocation mechanism compared to the LoRA baseline. Peak VRAM and Params represent the maximum memory usage and the number of trainable parameters, respectively.}
  \label{tab:appendix-efficiency}
\end{table}

We further analyze the computational and memory efficiency of different adaptive rank allocation methods on Mistral-7B. The comparison in Table~\ref{tab:appendix-efficiency} focuses on three practical aspects: additional allocation overhead, peak VRAM usage, and the number of trainable parameters. These factors are important because adaptive rank allocation is only useful in practice if the allocation procedure itself does not introduce prohibitive training cost.

Compared with AdaLoRA, RSRA avoids iterative rank pruning and training-time budget adjustment. As a result, it substantially reduces the extra allocation overhead. For example, on CoLA, the additional overhead of RSRA is only 62.3 seconds, whereas AdaLoRA requires 1620.8 seconds. This corresponds to more than a 26$\times$ reduction in extra overhead. A similar trend is observed on PIQA, where RSRA also achieves the lowest additional overhead among all adaptive methods.

Compared with SR-LoRA and GoRA, RSRA provides a better balance between efficiency and accuracy. Although SR-LoRA and GoRA also reduce the number of trainable parameters, their performance can degrade noticeably, especially for GoRA. In contrast, RSRA reduces the trainable parameter count by approximately 40\% compared with uniform LoRA, while still improving the final accuracy on both PIQA and CoLA. This indicates that the proposed representational probing mechanism does not merely compress the rank budget, but reallocates it toward modules that are more useful for downstream adaptation.

\begin{table*}[t]
  \centering
  \small
  \setlength{\tabcolsep}{5pt}
  \begin{tabular}{l ccccccc c}
    \toprule
    \textbf{Task}
    & $\mathbf{W_q}$
    & $\mathbf{W_k}$
    & $\mathbf{W_v}$
    & $\mathbf{W_o}$
    & $\mathbf{W_{\mathrm{gate}}}$
    & $\mathbf{W_{\mathrm{up}}}$
    & $\mathbf{W_{\mathrm{down}}}$
    & \textbf{Avg.} \\
    \midrule
    MRPC
    & 0.01449
    & 0.01178
    & 0.01630
    & 0.01745
    & 0.01700
    & 0.01805
    & 0.01032
    & 0.01506 \\

    RTE
    & 0.01290
    & 0.01080
    & 0.01617
    & 0.01690
    & 0.01521
    & 0.01662
    & 0.00953
    & 0.01402 \\

    WNLI
    & 0.00683
    & 0.00566
    & 0.00879
    & 0.00884
    & 0.00765
    & 0.00846
    & 0.00453
    & 0.00725 \\
    \midrule
    \textbf{Avg.}
    & 0.01141
    & 0.00941
    & 0.01375
    & 0.01440
    & 0.01329
    & 0.01438
    & 0.00813
    & \textbf{0.01211} \\
    \bottomrule
  \end{tabular}
  \caption{
  Relative magnitude of learned LoRA updates across tasks and module
  types. Each entry reports $\rho_{t,j}$, the relative Frobenius norm
  of the learned update averaged across Transformer layers. }
  \label{tab:epsilon_scale}
\end{table*}

In terms of memory usage, RSRA maintains a peak VRAM footprint close to that of the standard LoRA baseline. This is expected because the probing procedure is forward-only and does not require maintaining additional gradient statistics or dynamic rank-search states throughout training. Therefore, RSRA is particularly suitable for resource-constrained fine-tuning scenarios where both parameter efficiency and allocation overhead are important.

\subsection{Choice of the Probe Magnitude}
\label{sec:epsilon_justification}

To determine an appropriate magnitude for the virtual low-rank
updates, we examine the empirical scale of the updates learned by
standard LoRA. Let $\mathcal{T}$ denote the set of evaluated tasks,
$\mathcal{L}$ the set of Transformer layers, and $\mathcal{J}$ the set
of adapted module types. For task $t\in\mathcal{T}$, layer
$l\in\mathcal{L}$, and module type $j\in\mathcal{J}$, let
$W_{l,j}$ denote the corresponding pretrained weight matrix, and let
$A_{t,l,j}$ and $B_{t,l,j}$ denote the learned LoRA matrices. We define
the layer-averaged relative update magnitude for task $t$ and module
type $j$ as
\begin{equation}
    \rho_{t,j}
    =
    \mathbb{E}_{l\sim\operatorname{Unif}(\mathcal{L})}
    \left[
        \frac{
            \left\lVert A_{t,l,j}B_{t,l,j}\right\rVert_F
        }{
            \left\lVert W_{l,j}\right\rVert_F
        }
    \right]
    =
    \frac{1}{|\mathcal{L}|}
    \sum_{l\in\mathcal{L}}
    \frac{
        \left\lVert A_{t,l,j}B_{t,l,j}\right\rVert_F
    }{
        \left\lVert W_{l,j}\right\rVert_F
    }.
    \label{eq:empirical_lora_scale}
\end{equation}
Here, the index $j$ identifies a module type shared across layers,
such as the query, key, value, output, gate, up-projection, or
down-projection module.

Table~\ref{tab:epsilon_scale} reports the resulting statistics for
vanilla LoRA trained on MRPC, RTE, and WNLI. Across the 21
task--module combinations, the relative update magnitude ranges from
$0.00453$ to $0.01805$, and its overall average is $0.01211$.
The learned updates are therefore consistently on the order of
$10^{-2}$ relative to the corresponding pretrained weights.

Under the Frobenius normalization used by the virtual low-rank probe,
we have
\begin{equation}
    \frac{\lVert\Delta W_{l,j}\rVert_F}
         {\lVert W_{l,j}\rVert_F}
    =
    \epsilon.
\end{equation}
Setting $\epsilon=0.01$ therefore makes the probe magnitude equal to
$1\%$ of the Frobenius norm of the corresponding pretrained weight.
This value lies within the empirical range of trained LoRA updates and
is close to their overall average of $0.01211$. We use the same
$\epsilon=0.01$ for all tasks, layers, and module types, yielding a
representative yet conservative perturbation scale to avoid
complex tuning. Together with Frobenius
normalization, this fixed scale ensures that the measured sensitivity
differences are not merely caused by variations in the absolute
magnitudes of the probe updates.

\subsection{Detailed Analysis of Calibration Set Size}
\label{sec:calibration_size}

We examine the effect of the calibration set size $N$ on three GLUE tasks with Mistral-7B. We vary $N$ from 64 to 1024 and use $N=256$, the setting adopted in our main experiments, as the reference. Stability is evaluated from two complementary perspectives: the Spearman Rank Correlation Coefficient between the module-wise Fr\'{e}chet Distance scores ($\mathrm{SRCC}_{\mathrm{FD}}$) and the correlation between the resulting integer rank configurations ($\mathrm{SRCC}_{\mathrm{rank}}$). We additionally report the allocation time, including representation collection, module-wise forward probing, covariance estimation, and Fr\'{e}chet Distance computation.

\begin{table}[t]
\centering
\setlength{\tabcolsep}{6pt}
\begin{tabular}{l c c c c}
\toprule
\textbf{Task}
& $N$
& \textbf{Time (s)}
& $\mathrm{SRCC}_{\mathrm{rank}}$
& $\mathrm{SRCC}_{\mathrm{FD}}$ \\
\midrule
\multirow{5}{*}{MRPC}
    & 64   & 109.8 & 0.968 & 0.979 \\
    & 128  & 115.6 & 0.971 & 0.985 \\
    & $\mathbf{256}^{\dagger}$
            & \textbf{125.9} & \textbf{1.000} & \textbf{1.000} \\
    & 512  & 147.2 & 0.968 & 0.985 \\
    & 1024 & 190.3 & 0.971 & 0.981 \\
    \midrule
\multirow{5}{*}{RTE}
    & 64   & 110.6 & 0.955 & 0.979 \\
    & 128  & 115.4 & 0.959 & 0.981 \\
    & $\mathbf{256}^{\dagger}$
            & \textbf{125.9} & \textbf{1.000} & \textbf{1.000} \\
    & 512  & 147.1 & 0.965 & 0.977 \\
    & 1024 & 189.9 & 0.951 & 0.977 \\
    \midrule
\multirow{5}{*}{WNLI}
    & 64  & 109.0 & 0.944 & 0.968 \\
    & 128 & 114.2 & 0.944 & 0.976 \\
    & $\mathbf{256}^{\dagger}$
            & \textbf{124.7} & \textbf{1.000} & \textbf{1.000} \\
    & 512 & 145.6 & 0.938 & 0.966 \\
    & 635 & 155.9 & 0.943 & 0.972 \\
\bottomrule
\end{tabular}

\vspace{4pt}
\parbox{0.96\linewidth}{
\textit{Note:} $\dagger$ denotes the calibration size used in the main experiments. The WNLI training split contains 635 examples; therefore, its largest setting uses all available training samples.
}

\caption{
Effect of the calibration set size $N$ on allocation time and stability with Mistral-7B. Correlations $\mathrm{SRCC}_{\mathrm{rank}}$ and $\mathrm{SRCC}_{\mathrm{FD}}$ are computed against the reference $N=256$.
}
\label{tab:calibration_size}
\end{table}

As shown in Table~\ref{tab:calibration_size}, the estimated sensitivity structure remains highly consistent across calibration sizes. Excluding the $N=256$ reference rows, the Fr\'{e}chet Distance rankings attain an average $\mathrm{SRCC}_{\mathrm{FD}}$ of $0.977$, with all values ranging from $0.966$ to $0.985$. The resulting integer rank configurations also remain stable, achieving an average $\mathrm{SRCC}_{\mathrm{rank}}$ of $0.956$ and a minimum of $0.938$. The modest difference between the two correlations arises from projecting continuous sensitivity scores onto bounded integer ranks, where small score variations can alter rounding near allocation boundaries.

The allocation time increases smoothly with $N$ and remains nearly identical across tasks at the same calibration size, showing that the cost is governed primarily by the backbone and the number of processed samples. On MRPC, for example, increasing $N$ from 64 to 1024 raises the allocation time from 109.8 to 190.3 seconds. Thus, a $16\times$ increase in calibration samples produces only a $1.73\times$ increase in wall-clock time, reflecting the combination of a fixed model-dependent cost and a sample-dependent forward-computation cost.

We use $N=256$ in the main experiments as a balanced operating point between calibration coverage and allocation cost. It incorporates substantially more task-specific evidence than the smaller settings while requiring approximately 125 seconds, compared with 146--147 seconds at $N=512$ and approximately 190 seconds at $N=1024$. Meanwhile, the strong agreement of both smaller and larger settings with $N=256$ indicates that the selected configuration lies within a stable sensitivity-estimation regime.

\subsection{Detailed Analysis of Rank Boundary Ablation}
\label{sec:appendix_rank_boundary}

The rank boundaries $R_{\min}$ and $R_{\max}$ control the feasible allocation space of RSRA. Intuitively, $R_{\min}$ determines the minimum adaptation capacity assigned to each module, while $R_{\max}$ controls how much additional capacity can be concentrated on highly sensitive components. A desirable allocation strategy should be robust to these boundary choices while still benefiting from sufficient flexibility on challenging tasks. 

\begin{table}[t]
    \centering
    \begin{tabular}{l cccc}
        \toprule
        & \multicolumn{4}{c}{$R_{max}$} \\
        \cmidrule(lr){2-5}
        $R_{min}$ & 16 & 32 & 64 & 128 \\
        \midrule
        0 & 84.43 & 84.55 & 83.92 & 83.88 \\
        1 & 84.68 & 84.13 & 84.30 & \textbf{85.35} \\
        2 & 84.55 & 84.85 & 84.43 & 84.68 \\
        4 & \textbf{85.40} & 85.06 & 85.06 & 85.19 \\
        \bottomrule
    \end{tabular}
    \caption{Ablation of rank constraints on ARC-easy. Performance is relatively stable across various rank boundaries for simpler tasks.}
    \label{tab:arc_easy}
\end{table}

\begin{table}[t]
    \centering
    \begin{tabular}{l cccc}
        \toprule
        & \multicolumn{4}{c}{$R_{max}$} \\
        \cmidrule(lr){2-5}
        $R_{min}$ & 16 & 32 & 64 & 128 \\
        \midrule
        0 & 58.87 & 59.64 & 59.22 & 59.64 \\
        1 & 59.64 & 60.24 & 58.79 & 59.98 \\
        2 & 60.15 & \textbf{60.32} & 59.81 & 60.07 \\
        4 & \textbf{60.24} & 59.56 & 59.90 & 59.64 \\
        \bottomrule
    \end{tabular}
    \caption{Ablation of rank constraints on ARC-challenge. Higher rank floors ($R_{min}$) are crucial for complex reasoning tasks.}
    \label{tab:arc_chal}
\end{table}

As shown in Table~\ref{tab:arc_easy}, ARC-Easy is relatively insensitive to the choice of rank boundaries. Across all combinations of $R_{\min}$ and $R_{\max}$, the performance remains within a narrow range, suggesting that simpler reasoning tasks do not require highly specialized rank distributions. In this setting, even conservative rank budgets are sufficient to capture the necessary adaptation signals, and increasing the rank ceiling provides only marginal gains. 

In contrast, Table~\ref{tab:arc_chal} shows that ARC-Challenge benefits more clearly from a non-trivial rank floor and moderate rank headroom. When $R_{\max}=32$, increasing $R_{\min}$ from 0 to 2 improves the accuracy from 58.87\% to 60.32\%. This indicates that overly sparse allocations may underfit complex reasoning patterns, since some modules still require a minimum level of adaptation capacity even if their probing scores are relatively small. However, further increasing the rank ceiling beyond 32 does not consistently improve performance. For example, under $R_{\min}=2$, the performance slightly decreases when $R_{\max}$ is enlarged from 32 to 128. This suggests that excessive allocation flexibility may introduce redundant capacity, which can make the adaptation more prone to fitting task-specific noise rather than generalizable reasoning structures. 

Overall, these results suggest that RSRA is not overly sensitive to boundary hyperparameters on simpler tasks, while more challenging tasks benefit from a balanced configuration that avoids both overly sparse and overly permissive allocations. In our main experiments, we therefore adopt a moderate setting that maintains a non-zero rank floor and sufficient rank headroom, providing a practical trade-off between expressiveness, stability, and parameter efficiency.

\end{document}